\documentclass[letterpaper]{article} 
\usepackage{aaai25}  
\usepackage{times}  
\usepackage{helvet}  
\usepackage{courier}  
\usepackage[hyphens]{url}  
\usepackage{graphicx} 
\usepackage{natbib}  
\usepackage{caption} 
\usepackage{algorithm}
\usepackage{algorithmic}

\usepackage{newfloat}
\usepackage{listings}
\DeclareCaptionStyle{ruled}{labelfont=normalfont,labelsep=colon,strut=off} 
\floatstyle{ruled}
\newfloat{listing}{tb}{lst}{}
\floatname{listing}{Listing}
\usepackage{xspace}
\newcommand{\ourmethod}{{\fontfamily{lmtt}\selectfont \textbf{FastGLT}}\xspace}
\usepackage{enumitem}
\usepackage{amsmath}
\newtheorem{Def}{Definition}
\usepackage{dsfont}
\usepackage{amssymb}
\usepackage{booktabs}
\usepackage{multirow}
\usepackage{makecell}

\title{Fast Track to Winning Tickets: \\Repowering One-Shot Pruning for Graph Neural Networks}
\author{
    Yanwei Yue\equalcontrib,
    Guibin Zhang\equalcontrib\thanks{Project Lead, $^\ddagger$Corresponding Author},
    Haoran Yang,
    Dawei Cheng$^\ddagger$
}
\affiliations{
    Department of Computer Science and Technology, Tongji University\\

    \texttt{\{2053965, bin2003, 2010498, dcheng\}@tongji.edu.cn}
}

\usepackage{bibentry}

\begin{document}

\maketitle

\begin{abstract}
Graph Neural Networks (GNNs) demonstrate superior performance in various graph learning tasks, yet their wider real-world application is hindered by the computational overhead when applied to large-scale graphs. To address the issue, the Graph Lottery Hypothesis (GLT) has been proposed, advocating the identification of subgraphs and subnetworks, \textit{i.e.}, winning tickets, without compromising performance. The effectiveness of current GLT methods largely stems from the use of iterative magnitude pruning (IMP), which offers higher stability and better performance than one-shot pruning. However, identifying GLTs is highly computationally expensive, due to the iterative pruning and retraining required by IMP.
In this paper, we reevaluate the correlation between one-shot pruning and IMP: while one-shot tickets are suboptimal compared to IMP, they offer a \textit{fast track} to tickets with a stronger performance. We introduce a one-shot pruning and denoising framework to validate the efficacy of the \textit{fast track}. Compared to current IMP-based GLT methods, our framework achieves a double-win situation of graph lottery tickets with \textbf{higher sparsity} and \textbf{faster speeds}. Through extensive experiments across 4 backbones and 6 datasets, our method demonstrates $1.32\% - 45.62\%$ improvement in weight sparsity and a $7.49\% - 22.71\%$ increase in graph sparsity, along with a $1.7-44 \times$ speedup over IMP-based methods and $95.3\%-98.6\%$ MAC savings. The source code is available at \url{https://github.com/yanweiyue/FastGLT}.
\end{abstract}

%

\section{Introduction}

Graph Neural Networks (GNN) \cite{kipf2016semi,hamilton2017inductive} have recently become the predominant approaches for various graph-related learning challenges, including node classification \cite{velickovic2017graph,cheng2023anti,wang2023snowflake,wang2024heterophilic}, link prediction \cite{zhang2018link,zhang2019inductive}, and graph classification \cite{ying2018hierarchical,zhang2018end,fang2024regularization}. 
Nonetheless, the significant computational challenges primarily arise from the over-parameterized GNN weights that are equipped with dense connections, as well as from the large-scale graph samples as input. These factors impede efficient feature aggregation during the training and inference processes of GNNs \cite{jin2021graph,zhang2024adaglt}. Worse still, these intrinsic limitations curtail the application of GNNs in large-scale scenarios, particularly under resource-restricted conditions.

\begin{figure}[!t]
\centering
\includegraphics[width=1\columnwidth]{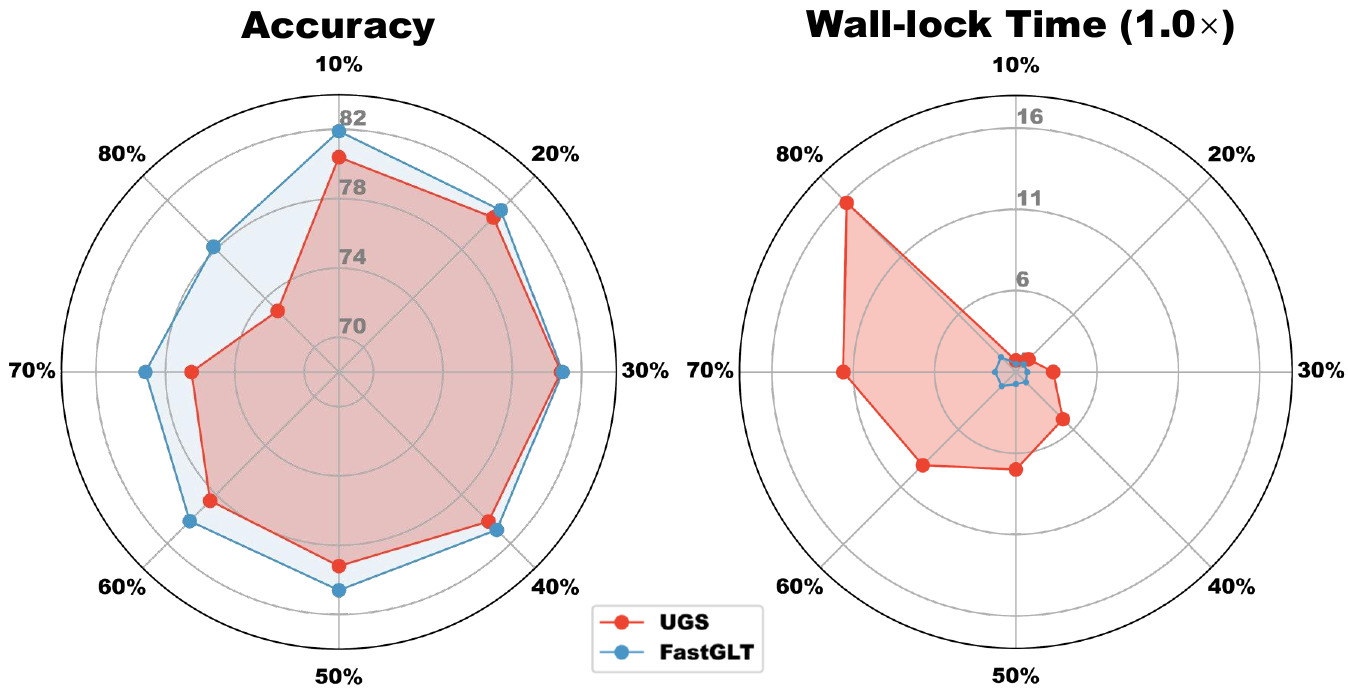}
\caption{(\textbf{\textit{Left}}) Accuracy ($\uparrow$) of UGS and \ourmethod on Cora+GAT, with fixed weight sparsity $s_\theta = 90\%$ and graph sparsity $s_g \in \{10\%, 20\%, \cdots, 80\%\}$ (\textbf{\textit{Right}}) Relative wall-clock time ($\downarrow$) compared to a single baseline training for searching GLTs. Note that \ourmethod requires far less wall-lock time to obtain subnetwork/subgraph with better performance than multiple rounds of IMP employed in UGS.}
\label{fig:intro}
\end{figure}

Prudently reflecting on prior research, the majority of efforts to tackle this inefficiency have concentrated on either (1) \textit{reducing the network's parameters} or (2) \textit{sparsifying the input graph}~\cite{zhang2024mog,chen2018fastgcn,zhang2024two,zhang2024adaglt}. The first class typically employs methods like quantization \cite{tailor2020degree}, pruning \cite{zhou2021accelerating}, and distillation \cite{chen2020self} to streamline GNN parameters. 
The second category involves leveraging graph sampling or sparsification techniques to reduce the computational demands caused by dense graphs~\cite{chen2018fastgcn,eden2018provable,li2020sgcn}. 
Recently, the Graph Lottery Ticket hypothesis (GLT) \cite{chen2021unified} takes the first step to unify the above two research lines. Briefly put, GLT aims to identify a \emph{graph lottery ticket}, \textit{i.e.}, a combination of \emph{a core subgraph and a sparse subnetwork} with admirable performance for accelerating GNN training and inference process. Such a hypothesis is inspired by the Lottery Ticket Hypothesis (LTH) \cite{frankle2018lottery}, which posits that sparse but performant subnetworks exist in a dense network with random initialization, like \emph{winning tickets} in a lottery pool. Considering its exceptional potential,
subsequent research has developed GLT from both theoretical~\cite{bai2022dual,wang2023brave} and algorithmic perspective~\cite{zhang2024adaglt,wang2023adversarial,wang2022pruning,you2022early}.

As one of the most well-established graph and weight pruning approaches, the success of GLT is attributed to the application of IMP, whose fundamental concept is to involve {iteratively pruning and retraining the model, methodically eliminating a small percentage of the remaining weights in each cycle, and persisting until the target pruning ratio is achieved.} Unfortunately, \textbf{the computational cost of IMP becomes excessively high as the targeted pruning ratio rises and pruned graph volume grows \cite{you2022early,zhang2021efficient},} as shown in Fig.~\ref{fig:intro} (\textit{Right}). To decrease computational expenses, several efficient one-shot magnitude pruning (OMP) methods have been introduced~\cite{lee2018snip,wang2020picking,ma2021sanity}, which directly prune the model to the desired sparsity level. However, they (1) typically exhibit a notable performance degradation compared to IMP~\cite{ma2021sanity,frankle2020pruning} and (2) mainly focus on weight pruning, and their performance in the context of joint pruning (graph/GNN) within the GLT context remains unexplored and unknown.

In this work, we take the \underline{first step} to explore the feasibility of utilizing one-shot pruning in place of IMP within the GLT context, aiming to break the persistent challenge that the performance of one-shot pruning has been inferior to IMP. Towards this end, we introduce a One-shot Pruning and Denoising Framework toward \underline{Fast} Track \underline{G}raph \underline{L}ottery \underline{T}ickets (termed \ourmethod). Technically, \ourmethod initially obtains tickets at a sparsity level close to the target through one-shot pruning, followed by denoising these tickets based on gradient and degree metrics to achieve performance comparable to traditional GLT derived from IMP. Our rationale for this approach stems from a straightforward motivation: Although subnetworks/subgraphs revealed by one-shot pruning are less optimal than those from IMP, the gap between these suboptimal tickets and IMP's winning tickets is minimal and exhibits consistent patterns.  Therefore, these one-shot tickets represent a \textit{fast track} to winning tickets. By denoising them, we can swiftly locate GLTs with significantly lower computational costs than those with IMP (shown in Fig.~\ref{fig:intro}). Our contributions can be summarized as follows:

\begin{itemize}[leftmargin=*]
    \item We re-evaluate one-shot pruning within the context of graph lottery tickets, hypothesizing and empirically validating a \textit{fast track} whereby one-shot tickets directly lead to high-performing winning tickets.
    \item We introduce a one-shot pruning and denoising framework (\ourmethod) for efficiently identifying GLTs. \ourmethod forgoes the expensive IMP steps in traditional ones, leveraging one-shot tickets as a fast track toward winning tickets accompanied by performance that is in no way inferior to that of IMP. 
    \item Extensive experiments on 6 datasets and 4 GNN architectures show that (i) \ourmethod achieves significant improvements in both weight sparsity ($5.82\% - 25.48\%\uparrow$) and graph sparsity ($3.65\% - 17.48\%\uparrow$) compared to current state-of-the-art GLT methods~\cite{chen2021unified,wang2023adversarial}, and (ii) \ourmethod demonstrates substantial efficiency, achieving a 1.7-44$\times$ speedup over IMP-based GLTs and $95.3\%-98.6\%$ MAC savings.
\end{itemize}

\section{Preliminary \& Motivation}

\subsection{Notations} 
We consider an undirected graph $\mathcal{G}=\{\mathcal{V},\mathcal{E}\}$, with $\mathcal{V}$ as the node set and $\mathcal{E}$ the edge set of $\mathcal{G}$. The feature matrix of $\mathcal{G}$ is represented as $\mathbf{X} \in \mathds{R}^{N \times F}$, where $N = |\mathcal{V}|$ signifies the total number of nodes in the graph. The feature vector for each node $v_i \in \mathcal{V}$, with $F$ dimensions, is denoted by $x_i = \mathbf{X}[i,\cdot]$.  An adjacency matrix $\mathbf{A} \in \{0,1\}^{N \times N}$ is utilized to depict the inter-node connectivity, where $\mathbf{A}[i,j] = 1$ indicates an edge $e_{ij} \in \mathcal{E}$, and $0$ otherwise.  Let $f(\cdot; \mathbf{\Theta})$ represent a GNN model with $\mathbf{\Theta}$ as its parameters. For instance, a two-layer GCN is formulated as:
\begin{equation}
\mathbf{Z} = f(\{\mathbf{A}, \mathbf{X}\}; \mathbf{\Theta}) = \text{Softmax}(\hat{A}\sigma(\hat{\mathbf{A}}\mathbf{X}\mathbf{\Theta}^{(0)})\mathbf{\Theta}^{(1)}),
\end{equation}
where $\mathbf{Z}$ denotes the output, $\hat{\mathbf{A}} = \hat{\mathbf{D}}^{-\frac{1}{2}}(\mathbf{A} + \mathbf{I}_n)\hat{\mathbf{D}}^{-\frac{1}{2}}$ is the normalized adjacency matrix, $\Tilde{\mathbf{A}} = \mathbf{A} + \mathbf{I}_n$, $\hat{\mathbf{D}}$ is the degree matrix of $\hat{\mathbf{A}}$, $\sigma(\cdot)$ is an activation function, and $\mathbf{\Theta}^{(k)}$ is the weight matrix at the $k$-th layer.

\subsection{Graph Lottery Ticket} 
Given an input graph $\mathcal{G}$ and a GNN model $f(\cdot; \mathbf{\Theta})$, let $\mathcal{G}_{\text{sub}} = \{\mathbf{A} \odot \mathbf{M}_g, \mathbf{X}\}$ be a subgraph of $\mathcal{G}$ and $f_{\text{sub}}(\cdot; \mathbf{\Theta} \odot \mathbf{M}_\theta)$ be a subnetwork of $f(\cdot; \mathbf{\Theta})$. Here, $\mathbf{M}_g$ and $\mathbf{M}_\theta$ are binary mask matrices for the adjacency matrix and model weights, respectively. Additionally, we can define the graph sparsity (GS) $s_g$ and weight sparsity (WS) $s_\theta$ as follows:
\begin{equation}
s_g = 1 - \frac{||\mathbf{M}_g ||_0}{ ||\mathbf{A} ||_0 }, \quad s_\theta = 1 - \frac{||\mathbf{M}_\theta||_0}{||\mathbf{\Theta} ||_0},
\end{equation}
where the $||\cdot||_0$ denotes the $\ell_0$ norm that counts the number of non-zero elements. The graph lottery ticket (GLT) is defined with $\mathcal{G}_{\text{sub}}$ and $f_{\text{sub}}$ as follows:
\begin{Def}
$\mathrm{(Graph\;Lottery\;Ticket)}.$ Let $\mathcal{G}$ represent an input graph, and let $f(\cdot; \mathbf{\Theta})$ denote a GNN with model parameters initialized at $\mathbf{\Theta}_{0}$. We define a graph lottery ticket as the pair $(\mathcal{G}_{sub}, f_{sub})$, where $\mathcal{G}_{sub}$ is a sparsified version of $\mathcal{G}$ and $f_{sub}$ corresponds to a sparsified model. This ticket satisfies the condition that, when trained in isolation, the performance metric $\varphi(f_{sub}(\mathcal{G}_{sub}; \mathbf{\Theta}_{0}))$ is at least $\varphi(f(\mathcal{G}; \mathbf{\Theta}_{0}))$, where $\varphi$ denotes the test accuracy.
\end{Def}
 We define the \emph{extreme graph/weight sparsity} of a GLT method as the maximum graph/weight sparsity where it successfully identifies GLTs.

\subsection{Motivation}\label{sec:motivation}
\begin{figure}[!t]
\centering
\includegraphics[width=1\columnwidth]{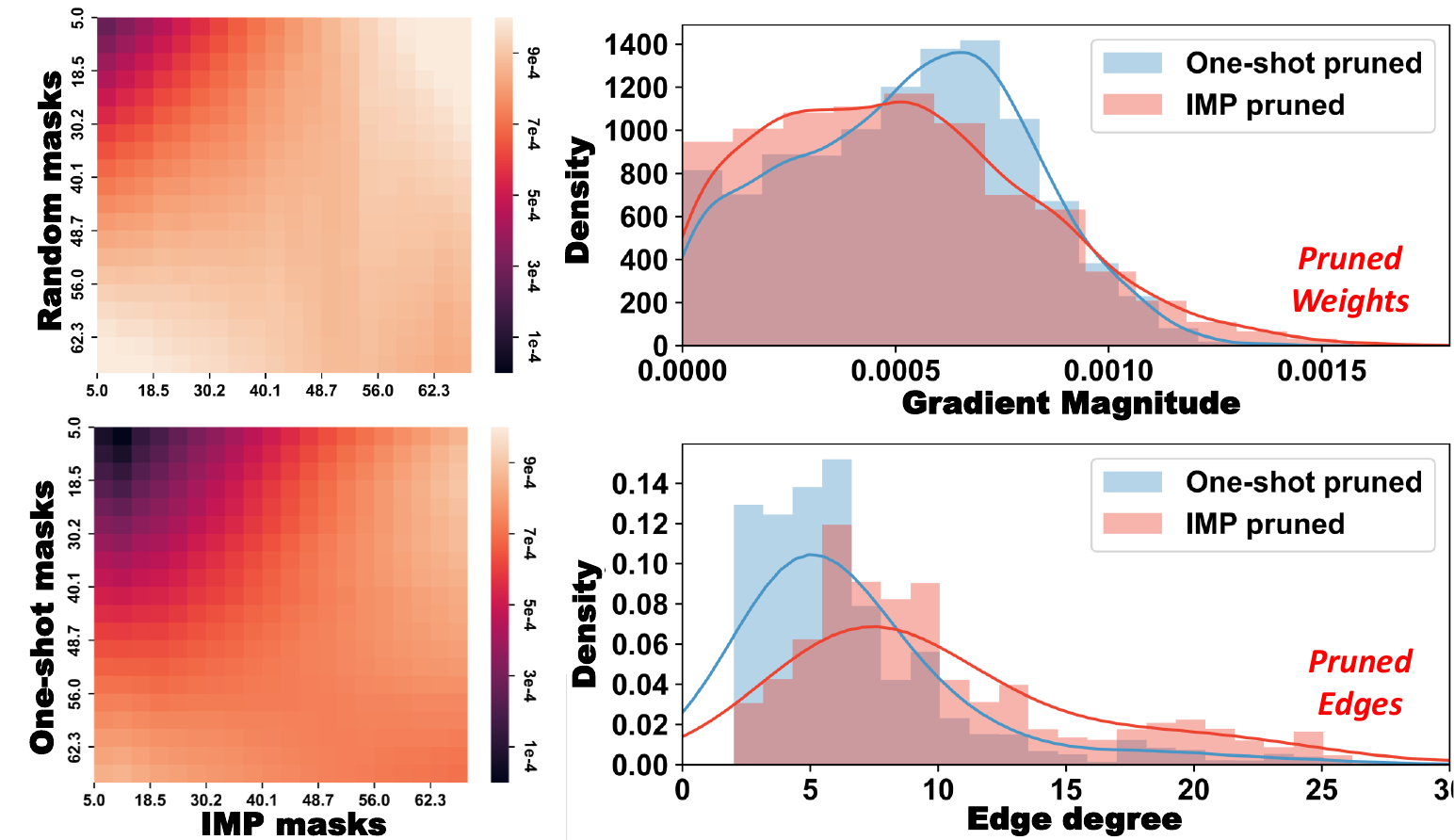}
\vspace{-2em}
\caption{(\textbf{\emph{Left}}) Hamming distance between masks generated by IMP, one-shot, and random sparsification methods on Cora with various graph sparsity levels $s_g \in \{5.0\%, 9.8\%, \ldots, 64.2\%\}$. Notably, as sparsity increases, the distance between random and IMP masks rapidly grows, whereas one-shot masks retain greater similarity to IMP masks. (\textbf{\emph{Right}}) Comparison of gradient magnitude/edge degree for weights/edges pruned by IMP or one-shot pruning.
}
\vspace{-1em}
\label{fig:motiv}
\end{figure}

\subsubsection{Comparison \& Visualization.} 
We define the graph mask $\mathbf{M}_g$ produced by UGS~\cite{chen2021unified} through iterative magnitude pruning as \emph{IMP-based masks}, those pruned by UGS directly to the target sparsity as \emph{one-shot masks}, and masks from random pruning as \emph{random masks}. Fig.~\ref{fig:motiv} (\emph{Left}) illustrates the Hamming distance~\cite{you2022early} between these masks at different sparsity levels, which can effectively reflect their differences. 
It is noticeable that the disparity between random and IMP masks (termed structural noise) snowballs with increasing sparsity. In contrast, the noise between one-shot and IMP masks consistently remains minimal.
This prompts us to consider: \emph{What characteristics define these structural noises?} 

\subsubsection{Emperical Validation.} 
Further, we employed gradients~\cite{evci2020rigging,lee2018snip} and edge degrees\footnote{For $e_{ij}$, we define its edge degree as $(|\mathcal{N}(v_i)| + |\mathcal{N}(v_j)|)/2$.}~\cite{wang2022pruning} to visualize differences in $\mathbf{M}_\theta$ and $\mathbf{M}_g$ between one-shot and IMP tickets, as shown in Fig.~\ref{fig:motiv} (\textit{Right}). Observations reveal that (1) weights pruned by IMP exhibit generally smaller gradients than those pruned one-shot, (2) degrees of edges pruned by IMP are significantly lower than those pruned one-shot. We draw an intuitive conclusion that, compared to IMP tickets, one-shot pruning's suboptimal performance stems from mistakenly pruning a minority of weights with higher gradients and edges with lower edge degrees. A natural question arises: \emph{can we enhance one-shot tickets' performance by denoising structural noise using gradient- and degree-based metrics?} This will be empirically validated in the following sections.

\section{Methodology}
\label{sec:methodology}

\begin{figure*}[!ht]
\centering
\includegraphics[width=0.9\textwidth]{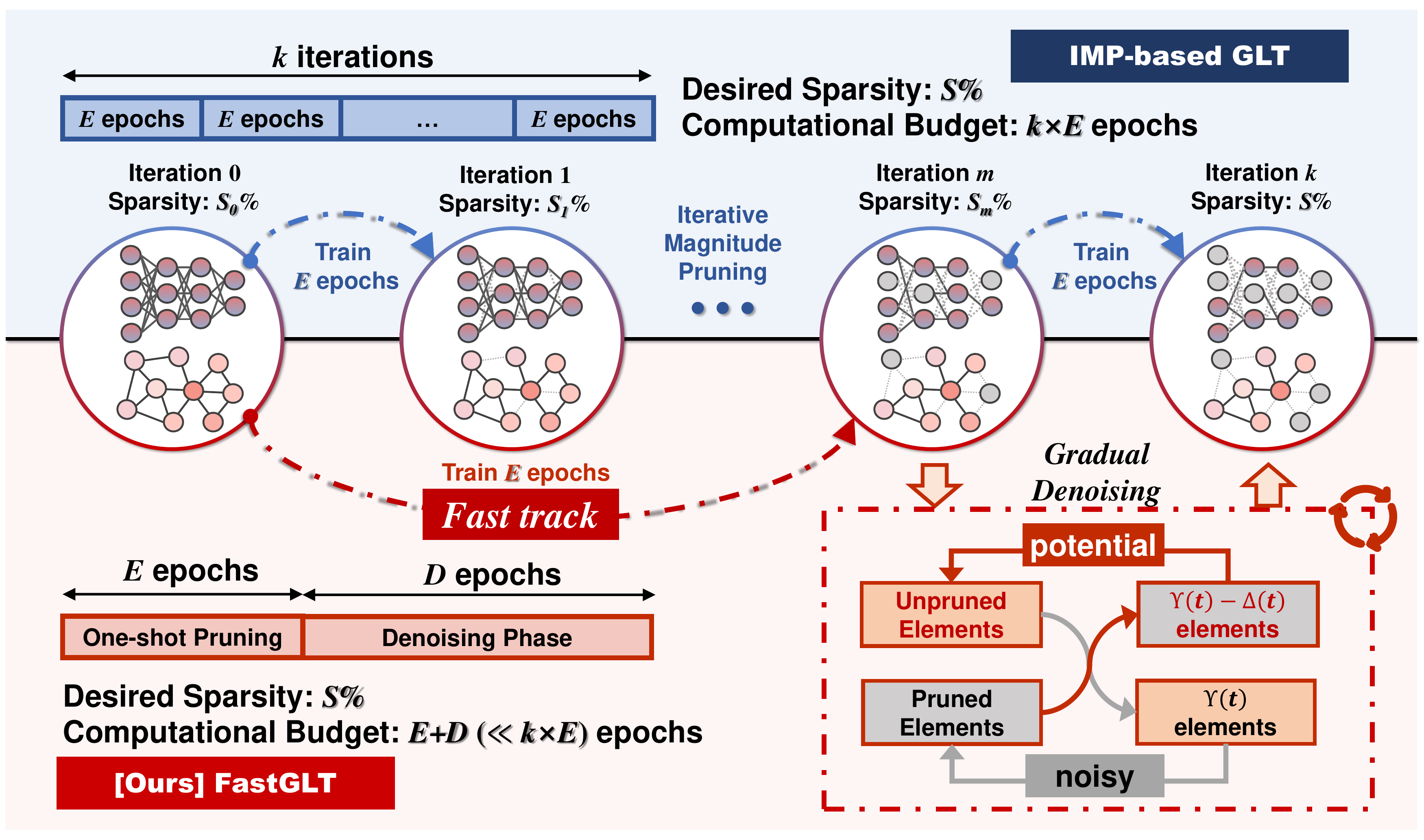}
\vspace{-1em}
\caption{The detailed illustration of  \ourmethod compared to conventional IMP-based GLT. \ourmethod replaces most of the time-consuming iterative stages with one-shot pruning as a \textit{fast track}, and leverages a gradual denoising module to fine-tune the one-shot tickets to the target sparsity with performance in no way inferior to that of IMP. 
}
\label{fig:framework}
\vspace{-1em}
\end{figure*}

Fig.~\ref{fig:framework} illustrates the comparison between our \ourmethod and IMP-based GLT methods like UGS and WD-GLT~\cite{hui2022rethinking}. Fig.~\ref{fig:framework} ({\textit{Up}}) depicts how traditional methods iteratively prune and retrain through $k$ iterations (each taking $E$ epochs) to achieve a GLT at target sparsity $S\%$. Conversely, Fig.~\ref{fig:framework} ({\textit{Down}}) shows that \ourmethod employs a one-shot pruning fast track to closely approach the target sparsity, followed by a gradual denoising process to fine-tune one-shot tickets towards better performance. This approach requires a total computational budget of $E+D$ epochs, significantly less than IMP's $k \times E$. 
In subsequent subsections, we outline in Sec.~\ref{sec:oneshot} how \ourmethod acquires one-shot tickets, and then elaborate on how the gradual denoising mechanism refines these tickets in Sec.~\ref{sec:denoise}.

\subsection{One-shot Pruning Tickets}
\label{sec:oneshot}
As depicted above, we assign two trainable masks $\boldsymbol{m}_g$ and $\boldsymbol{m}_\theta$ on input graph $\mathcal{G}$ and GNN model $f(\cdot; \mathbf{\Theta})$ (initialized with $\mathbf{\Theta}_0$).
Firstly, we co-optimize $\mathbf{A}$, $\mathbf{\Theta}$, $\boldsymbol{m}_g$, and $\boldsymbol{m}_\theta$ in an end-to-end manner using the following objective function:
\begin{equation}\label{eq:loss}
\mathcal{L}_{\text{os}} = \mathcal{L} \left( f_{sub}\left( \{\boldsymbol{m}_g \odot \mathbf{A}, \mathbf{X} \}, \boldsymbol{m}_\theta \odot \mathbf{\Theta}\right) ;y \right),
\end{equation}
where $\mathcal{L}$ denotes the task-itrelevant loss function (\textbf{i.e.}, cross-entropy loss), $\odot$ denotes element-wise multiplication and $y$ denotes the node labels. Different from UGS~\cite{chen2021unified}, we do not impose $\ell_1$ regularization on $\boldsymbol{m}_g$ and $\boldsymbol{m}_\theta$ to reduce the usage of hyperparameters. Upon completing the training, we select the optimal masks, \textit{i.e.}, $\boldsymbol{m}_g$ and $\boldsymbol{m}_\theta$ from the epoch with the highest validation score, denoted as $\boldsymbol{m}^*_g$ and $\boldsymbol{m}^*_\theta$, for pruning. 

Given the target graph sparsity $s^{\text{tgt}}_g$ and weight sparsity $s^{\text{tgt}}_\theta$, we avoid pruning directly to these target sparsities. This is due to the potential performance collapse~\cite{hui2022rethinking} when targeting extremely high sparsity (\textit{e.g.}, 99\% weight sparsity or 80\% graph sparsity), which can render the model untrainable. Instead, we utilize an exponential decay function $\mathrm{\Psi}(s) = s - \alpha s^{\beta}$ to pre-calculate an intermediate sparsity $s^{\text{inm}}_g$ and $s^{\text{inm}}_\theta$ based on the target sparsities, and $\alpha$ and $\beta$ are coefficients to adjust the output. We then zero the lowest-magnitude elements in $\boldsymbol{m}^*_g$ and $\boldsymbol{m}^*_\theta$ w.r.t. $s^{\text{inm}}_g$ and $s^{\text{inm}}_\theta$ in a unified manner, as outlined below:
\begin{equation}
\mathbf{M}^{\circledcirc} = \mathds{1}\left[\boldsymbol{m}^*\right] \odot \mathds{1}\left[|\boldsymbol{m}^*| > \operatorname{Thresholding} (\boldsymbol{m}^*, \mathrm{\Psi}(s^{\text{tgt}}))\right],
\end{equation}
where
$\mathbf{M}^{\circledcirc} \in \{0,1\}^{|\boldsymbol{m}^*|}$ is either the one-shot graph mask $\mathbf{M}^{\circledcirc}_g$ or weight mask $\mathbf{M}^{\circledcirc}_\theta$, $\mathds{1}[\cdot]$ is a binary indicator, and $\operatorname{Thresholding}(m, s)$ denotes caculating the global threshold value at top $s$ by sorting $m$ in descending order.

\subsection{Gradual Denoising Mechanism}
\label{sec:denoise}

As stated in~\cite{wang2023adversarial}, traditional GLT methods irreversibly exclude elements pruned in a given iteration from subsequent considerations, leading to information loss in the pruned subgraph/subnetwork. This aligns with our assertion: one-shot masks may contain noisy and ineffective elements, while pruned parts could hold valuable structures. Towards this end, we proposed a gradual denoising mechanism, which repeatedly identifies the noisy elements in the current subgraph/subnetwork and replaces them with potential ones in the pruned components within $D$ epochs.

\noindent \textbf{Noisy Component Identification.} Given the one-shot masks $\mathbf{M}^{\circledcirc}_g$ and $\mathbf{M}^{\circledcirc}_\theta$, 
we train the model with fixed sparse masks and a trainable graph mask $\boldsymbol{m}_g$, denoted as $f\left(  \{\boldsymbol{m}_g \odot \mathbf{M}^{\circledcirc}_g \odot \mathbf{A}, \mathbf{X}\},  \mathbf{M}^{\circledcirc}_\theta \odot \mathbf{\Theta} \right)$ with the objective function similar to Eq.~\ref{eq:loss}. 
We execute progressive denoising over intervals spanning $\Delta T$ epochs, and the $d$-th epoch is therefore assigned to the $\lceil d / \Delta T \rceil^{th}$ interval.
At the end of interval $\mu (1 \leq \mu \leq \lceil D / \Delta T \rceil = \mu^{\text{end}})$, analogous to traditional magnitude pruning, elements with the smallest magnitudes after $\Delta T$ training epochs are considered noisy components, as defined as follows:

\begin{equation}\label{eq:noisy}
\begin{cases}
\mathbf{M}_\theta^{(\text{ns})} = \mathcal{F}\left(-|\mathbf{M}_\theta^{(\mu)} \odot \mathbf{\Theta}_{\Delta T}|,\; \mathbf{N}^{(\text{ns})}_\theta \right) \\

\mathbf{M}_g^{(\text{ns})} = \mathcal{F}\left(-|\mathbf{M}_g^{(\mu)}\odot\boldsymbol{m}_g|, \;\mathbf{N}^{(\text{ns})}_g \right)\\ 
    \end{cases}   
\end{equation}

where $\mathbf{M}_\theta^{(\text{ns})}$ and $\mathbf{M}_g^{(\text{ns})}$ are identified noisy weight/edges, $\mathbf{N}^{(\text{ns})}_\theta = \#\mathbf{M}^{(\mu)}_\theta\times\mathbf{\Upsilon}(\mu)$ and $\mathbf{N}^{(\text{ns})}_g = \#\mathbf{M}^{(\mu)}_g\times\mathbf{\Upsilon}(\mu)$ are the number of identified noisy weight/edges, $\mathcal{F}(m,k)$ returns the indices of top-$k$ elements of matrix $m$, and $\#$ counts the number of elements in a matrix. $\mathbf{\Upsilon}(\cdot)$ is a denoising scheduler that outputs the ratio of weights/edges to be identified between each interval. Here, we adopt the Inverse Power~\cite{zhu2017prune,evci2020rigging}, $\mathbf{\Upsilon}(\mu) = \tau (1 - \mu/\mu^{\text{end}})^\kappa$, where $\tau$ denotes the initial ratio and $\kappa$ is the decay factor controlling how fast the ratio decreases with intervals.

\noindent \textbf{Potential Component Discovery.} Slightly differently, in unearthing potentially important weights/edges, we adopt new metrics based on our observations in Sec.~\ref{sec:motivation}. 
Technically, for pruned weights $\neg \mathbf{M}_\theta^{(\mu)}\odot\mathbf{\Theta}$, we identify those with the highest accumulated gradients as potential ones. For pruned edges $\neg\mathbf{M}^{(\mu)}_g\odot\mathbf{A}$, we regard those with the smallest edge degrees as potential.
Notably, to gradually increase the graph/weight sparsity from the initial $s_g^{\text{inm}}$ and $s_\theta^{\text{inm}}$ to the target $s_g^{\text{tgt}}$ and $s_\theta^{\text{tgt}}$, we ensure that the identified important elements are fewer than the noisy ones by a factor of $\omega_g=\frac{\#||{A}||_0 \times (s_g^{\text{tgt}} - s_g^{\text{inm}})}{\mu^{\text{end}}}\%$ or $\omega_\theta=\frac{\#\mathbf{\Theta} \times (s_g^{\text{tgt}} - s_g^{\text{inm}})}{\mu^{\text{end}}}\%$ between each interval. The process is defined as follows:
\begin{equation}\label{eq:potential}
\small
\begin{cases}
\mathbf{M}_\theta^{(\text{pt})} = \mathcal{F}\left(\sum\limits_{i=1}^{\Delta T}\left|\nabla_{\neg \mathbf{M}_\theta^{(\mu)}\odot\mathbf{\Theta}}\mathcal{L}  \right|,\; \mathbf{N}_\theta^{(\text{pt})} \right) \\

\mathbf{M}_g^{(\text{pt})} = \mathcal{F}\left(-(\neg\mathbf{M}^{(\mu)}_g\odot\mathbf{A}\odot\mathbf{S}^{(\mu)}),\; \mathbf{N}_g^{(\text{pt})}\right),\\
    \end{cases}
\end{equation}
where $\mathbf{M}_\theta^{(\text{pt})}$ and $\mathbf{M}_g^{(\text{pt})}$ are potentially important weight/edges discovered from pruned elements, $\mathbf{N}_\theta^{(\text{pt})} = \#\mathbf{M}^{(\mu)}_\theta\times \mathbf{\Upsilon}(t) - \omega_\theta $ and $\mathbf{N}_g^{(\text{pt})} = \#\mathbf{M}^{(\mu)}_g\times\mathbf{\Upsilon}(t) - \omega_g$ are the number of potentially important weight/edges, $\sum_{i=1}^{\Delta T}|\nabla_{\neg \mathbf{M}_\theta^{(\mu)}\odot\mathbf{\Theta}}\mathcal{L}  |$ calculates the accumulated gradients of pruned weights in interval $\mu$, and $\mathbf{S}$ is the edge degree matrix, calculated as follows:
\begin{equation}\label{eq:update}
\mathbf{S} = (\mathbf{D}^{-1} \mathbf{M}_g^{(\mu)} \operatorname{d}(\mathbf{D})^{-\frac{1}{2}})(\mathbf{\mathbf{D}}^{-1} \mathbf{M}_g^{(\mu)} \operatorname{d}(\mathbf{D})^{-\frac{1}{2}})^\text{T},
\end{equation}
where $\mathbf{D}$ denotes the degree matrix of the sparsed graph $\mathbf{M}^{(\mu)}_g$ and $\operatorname{d(\mathbf{D})}$ returns the node degree vector of $\mathcal{V}$.

\noindent \textbf{Mask Update.} 
Now that we have identified both noisy and potential edges/weights, we proceed to update the graph/weight masks. Specifically, we remove the noisy components from the current mask and incorporate the potential ones, as detailed in the following process:
\begin{equation}
\mathbf{M}^{(\mu+1)} = \left( \mathbf{M}^{(\mu)} \setminus \mathbf{M}^{(\text{ns})} \right) \cup \mathbf{M}^{(\text{pt})},
\end{equation}
where $\mathbf{M}^{(\mu+1)}$ is either the updated graph mask $\mathbf{M}_g^{(\mu+1)}$ or weight mask $\mathbf{M}_\theta^{(\mu+1)}$. Note that between each interval, the sparsity of $\mathbf{M}^{(\mu+1)}$ increases by $\left(\frac{s^{\text{tgt}} - s^{\text{inm}}}{\mu_{\text{end}}}\right)\%$ compared to $\mathbf{M}^{(\mu)}$, ensuring that the graph or network precisely reaches the target sparsity at the end of the denoising process.

During the continuous identifying and swapping process, both the network and the graph are denoised to the desired sparsity with satisfactory performance. The overall algorithm framework is showcased in Algo.~\ref{alg:algo}.

\section{Experiments}

\begin{figure*}[!t]
\centering
\includegraphics[width=1\textwidth]{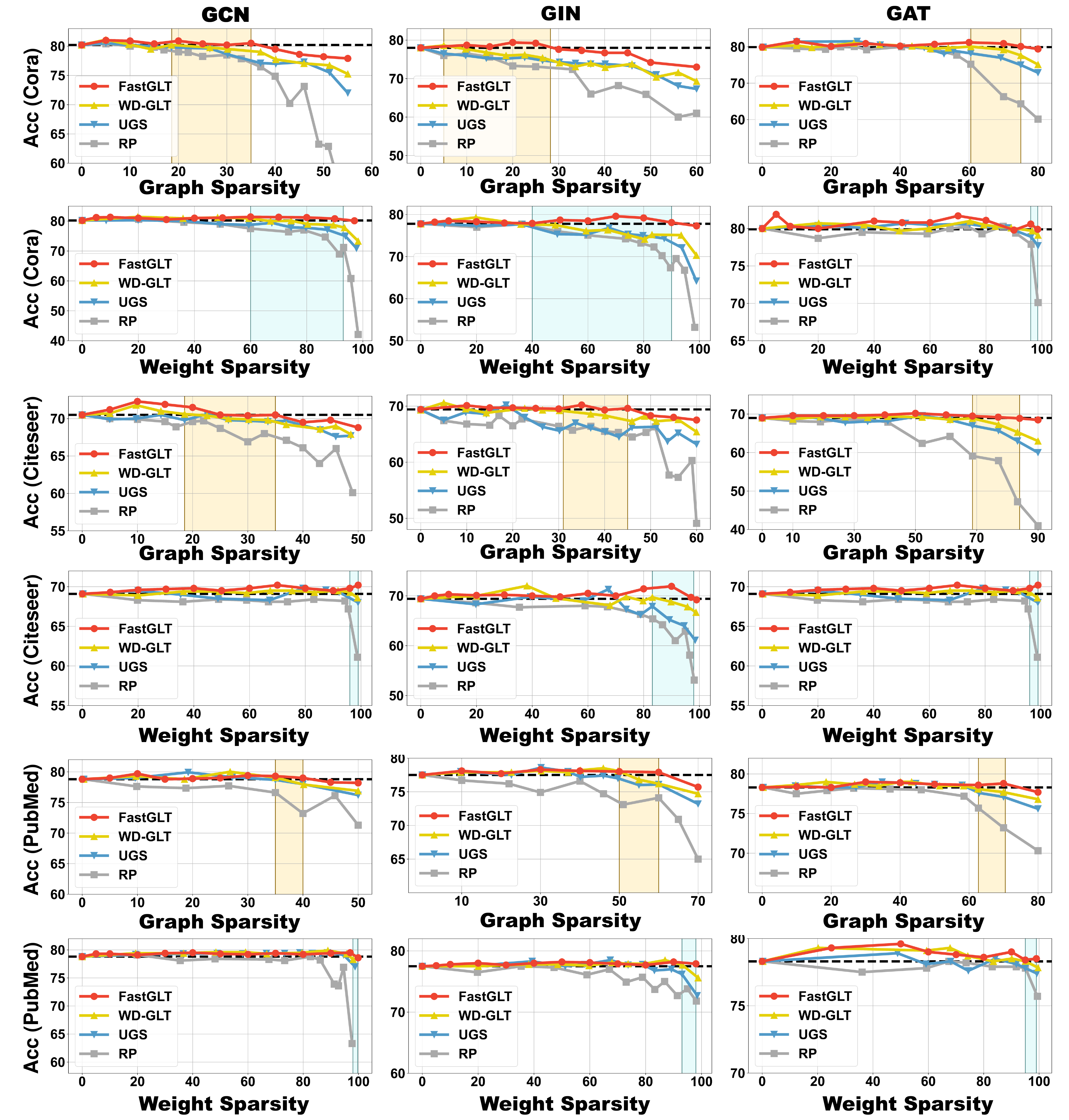}
\vspace{-1.5em}
\caption{Results of node classification over Cora/Citeseer/PubMed with GCN/GIN/GAT backbones. Black dash lines represent the baseline performance.}
\label{fig:rq1}
\vspace{-1em}
\end{figure*}

In this section, we conduct extensive experiments to answer the following three research questions: (\textbf{RQ1}) Can \ourmethod effectively find graph lottery tickets? (\textbf{RQ2}) Can \ourmethod scale up to larger-scale graphs? (\textbf{RQ3}) Does \ourmethod genuinely accelerate the acquisition of winning tickets and inference speed compared to traditional IMP-based GLT?

\begin{table*}[!t]
\centering
\scriptsize
\setlength{\tabcolsep}{5.40pt}
\caption{Efficiency comparison among UGS, WD-GLT and \ourmethod. ``Acc. (\%)'' indicates the accuracy of the sparsest winning tickets obtained; ``Obt. Time (s)'' represents the wall-clock time consumed to obtain the sparsest winning ticket (for baseline, it refers to the full training time with dense network/graph); ``Inference MACs (M)'' refers to the inference MACs ($=\frac{1}{2}$FLOPs) required by the identified tickets; ``Relative Time (s)'' refers to the time relative to the original dense training duration.}
\label{tab:efficiency}
\begin{tabular}{lcccccccccccc}
\toprule
\multirow{2}{*}{Model} & \multirow{2}{*}{Methods} & \multicolumn{3}{c}{Cora} & \multicolumn{3}{c}{Citeseer} & \multicolumn{3}{c}{PubMed}  & \multicolumn{2}{c}{Avg. } \\
\cmidrule(lr){3-5} \cmidrule(lr){6-8}\cmidrule(lr){9-11} \cmidrule(lr){12-13}
 & & \makecell{Acc. \\(\%)}& {\makecell{Obt.\\Time}} & \makecell{Inf.\\MACs } & \makecell{Acc. \\(\%)}& \makecell{Obt.\\Time } & \makecell{Inf.\\MACs} &\makecell{Acc. \\(\%)}& \makecell{Obt.\\Time } & \makecell{Inf.\\MACs } & \makecell{Relative\\Time } & \makecell{MAC\\Savings }\\
\midrule
GCN & Baseline & $80.25 \pm 0.18$ & 21.4 & 1996M & $70.51 \pm 0.06$ & 41.4 & 6317M & $78.80 \pm 0.14$ & 1217.3 & 5077M & $1.0\times$  & $0.0\%$ \\

 & UGS & $80.19 \pm 0.06$ & 76.3 & 817M & $70.66 \pm 0.08$ & 233.8 & 1665M & $79.01 \pm 0.23$ & 3500.9 & 233M & $4.0\times$ & $75.6\%$ \\
 & WD-GLT & $80.27 \pm 0.45$ & 604.3 & 806M & $70.74 \pm 0.51$ & 801.7 & 1541M  & $78.82 \pm 0.33$ & 5634.1 & 107M & $17.4\times$ & $80.3\%$ \\
 & \ourmethod & $80.33 \pm 0.17$ & 34.9 & 139M & $70.59 \pm 0.12$ & 89.7 & 315M & $79.11 \pm 0.29$ & 1366.2 & 50M & $1.63\times$& $95.6\%$ \\
\midrule

GIN & Baseline & $78.06 \pm 0.09$ & 7.3 &  2006M & $68.47 \pm 0.11$ & 8.6 &  6328M & $77.50 \pm 0.17$ & 8.8 &  5108M  &$1.0\times$& $0.0\%$  \\
 & UGS & $78.17 \pm 0.13$ & 39.8 & 1284M & $68.70 \pm 0.20$ & 61.0 & 2073M  & $77.66 \pm 0.30$ & 141.7 & 438M & $28.6\times$& $64.3\%$ \\
 & WD-GLT & $78.26 \pm 0.44$ & 305.4 & 1275M & $68.50 \pm 0.38$ & 891.4 & 1018M & $77.80 \pm 0.35$ & 1509.3& 331M & $105.6\times$ & $70.6\%$\\
 & \ourmethod & $78.32 \pm 0.17$ & 20.4 & 200M & $68.54 \pm 0.25$& 21.3 & 126M& $77.59 \pm 0.37$ & 17.8 & 102M & $2.4\times$ & $95.3\%$\\
 \midrule
 
GAT & Baseline & $79.95 \pm 0.03$ & 333.1 & 16059M & $69.12 \pm 0.18$ & 284.1 & 50619M & $78.35 \pm 0.20$ & 920.7 & 41349M & $1.0\times$  &$0.0\%$ \\
 & UGS & $79.99 \pm 0.45$ & 3143.8 & 1672M & $69.23 \pm 0.11$ & 3434.1 & 839M& $78.39 \pm 0.34$ & 2960.5 & 3565M & $8.2\times$ &$93.2\%$ \\
 & WD-GLT & $80.11 \pm 0.61$ & 5700.7 & 525M & $69.38 \pm 1.23$ & 5003.6 & 568M & $78.52 \pm 0.20$ & 4533.2 & 2283M & $13.2\times$ &$96.4\%$\\
 & \ourmethod & $80.07 \pm 0.37$ & 525.9 & 204M & $69.30 \pm 0.21$ & 528.7 & 414M  & $78.56 \pm 0.68$ & 1270.3 & 414M & $4.8\times$  &$98.6\%$\\
\bottomrule
\end{tabular}
\end{table*}

\subsection{Experiment Setup}

\noindent \textbf{Datasets.}
We select Cora, Citeseer, and PubMed~\cite{kipf2016semi} for node classification. For larger-scale graphs, we opt Ogbn-Arxiv/Proteins/Collab~\cite{hu2020open}. For a fair comparison, we follow the datasets splitting criterion used by UGS~\cite{chen2021unified}. On small-scale datasets, we use 140 (Cora), 120 (Citeseer), and 60 (PubMed) labeled nodes for training, 500 nodes for validation and 1000 nodes for testing. For OGB datasets, we follow the official splits given in~\cite{hu2020open}.

\noindent \textbf{Backbones \& Baselines.}
To assess \ourmethod's adaptability across various GNN backbones, we employ three network structures for small-scale datasets: GCN~\cite{kipf2017semi}, GIN~\cite{xu2018powerful} and GAT~\cite{velivckovic2017graph}. For larger-scale datasets, we utilize a 28-layer ResGCN~\cite{li2020deepergcn}. To comprehensively validate the efficiency of \ourmethod, we select two state-of-the-art GLT methods, UGS~\cite{chen2021unified} and WD-GLT~\cite{hui2022rethinking}, alongside random pruning (RP), for comparison. 

\noindent \textbf{Parameter Settings.}
For small-scale datasets, the hidden dimension is uniformly set to 512. On OGB graphs, we adopt parameter settings similar to UGS~\cite{chen2021unified}. Adam is used as the optimizer throughout. To compute intermediate sparsity, we employ $\mathrm{\Psi}(s)=s-0.01s^{1.2}$. The decay factor $\kappa$ is set as $1$ in all experiments. Detailed hyperparameter settings are provided in the Appendix.

\subsection{Results on Small Graphs ($\mathcal{RQ}$1)}
\label{sec:rq1}
To answer $\mathcal{RQ}$1, we compare our \ourmethod with UGS, WD-GLT and random pruning on three small-scale datasets for node classification tasks. Following~\cite{wang2023searching,wang2023adversarial}, toward a clearer illustration, we fix the weight sparsity to zero when investigating how accuracy evolves with the increase of graph sparsity, and vice versa. Fig.~\ref{fig:rq1} illustrates the results on Cora, Citeseer, and PubMed, and we can draw the following observations (\textbf{Obs}):

\noindent \textbf{Obs.1. \ourmethod can find GLTs with sparser subgraph/subnetwork.} It is observable that \ourmethod consistently outperforms other GLT methods across all backbones and benchmarks, attaining improvements in weight sparsity from 1.32\% to 45.62\% and in graph sparsity from 7.49\% to 22.71\%. Specifically, on Cora+GIN, the GLT identified by \ourmethod achieves 28.66\% graph sparsity and 89.16\% weight sparsity, surpassing WD-GLT by 22.71\% and 45.62\% respectively.

\noindent \textbf{Obs.2. \ourmethod demonstrates greater robustness in graph sparsification. } \ourmethod uniquely sustains performance with rising graph sparsity, in contrast to typical GLT methods. On Citeseer+GAT, for instance, while UGS and WD-GLT sharply decline in performance beyond 70\% graph sparsity, \ourmethod remains close to baseline at nearly 90\%.

\begin{table}[!t]\scriptsize
\centering
\caption{Results on 3 large-scale OGB graphs. 
Each entry denotes the extreme sparsity that a certain method is capable of achieving. \textit{Please note that extreme sparsity refers to the highest sparsity level at which GNN can achieve performance equal to the vanilla GNN.} $\pm$ corresponds to the standard deviation over 5 trials. ``N/A'' means GLT cannot be found.}
\setlength{\tabcolsep}{3.40pt}
\begin{tabular}{lccc}
\toprule
Dataset & Ogbn-Arxiv & Ogbn-Proteins & Ogbl-Collab  \\ 
\hline
\multicolumn{4}{c}{\textbf{Graph Sparsity}}  \\ \hline
Random Pruning  & N/A & $5.74\pm2.05$ &  N/A \\
UGS\citep{chen2021unified}   & $11.19\pm 0.42$ & $16.94\pm0.33 $  & $8.20\pm0.14$\\
WD-GLT\citep{hui2022rethinking} & $30.94\pm 0.51$ & $22.48\pm0.07$ & $17.14\pm0.95$ \\
\ourmethod (Ours)  & {{$48.01 \pm 0.17$}} & {{$34.49\pm 0.13$}} &  {{$31.55 \pm 0.35$}}\\
\bottomrule
\end{tabular}
\label{tab:large_graph}
\end{table}

\subsection{Results on Large Graphs ($\mathcal{RQ}$2)}
\label{sec:rq2}
To answer $\mathcal{RQ}$2, we conduct comparative experiments on Ogbn-Arxiv, Ogbn-Proteins and Ogbl-Collab with 28-layer ResGCN~\cite{li2020deepergcn}. As showcased in Tab.~\ref{tab:large_graph} and Tab.~\ref{tab:large_graph_weight}, we can list the following observations:

\noindent \textbf{Obs. 3. \ourmethod can scale up to large graphs.} \ourmethod consistently identifies GLTs with weight sparsity over 70\% and graph sparsity over 30\% across three datasets, surpassing other methods which generally fall below 60\% and 30\%, respectively. Specifically, \ourmethod can find GLT with 70.25\% weight sparsity or 48.01\% graph sparsity on Ogbn-Arxiv, exceeding UGS by 30.18\% and 36.82\%.

\begin{figure}[!t]
\centering
\includegraphics[width=1\columnwidth]{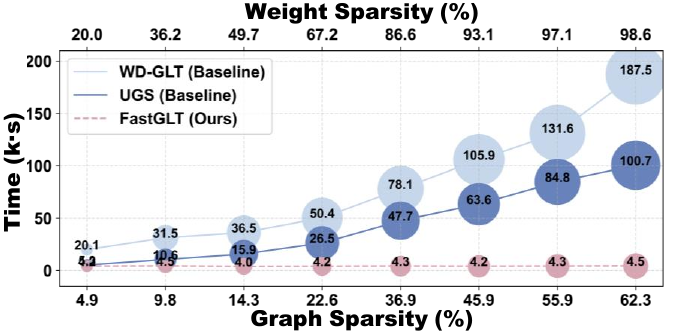}
\caption{The wall-lock time of UGS,
WD-GLT, FastGLT to locate GLTs
on Ogbn-Arxiv with different $s_g$ and $s_\theta$.}
\label{fig:time_arxiv}
\end{figure}

\subsection{Efficiency Validation ($\mathcal{RQ}$3)}
\label{sec:efficiency}

In this section, we compare the efficiency of \ourmethod with previous GLT methods from two perspectives: (1) wall-clock time expended in the search for winning tickets, and (2) inference speed of the most sparse tickets discovered. From Tab.~\ref{tab:efficiency} and Fig.~\ref{fig:time_arxiv}, we draw a conclusion that \ourmethod \textit{achieves a dual win in GLT search time and computational savings} from the following observations:

\noindent \textbf{Obs.4. \ourmethod can identify GLTs way faster. }
Specifically, UGS requires $4.0-28.6\times$ the duration of the original dense training to find the sparsest GLT. On GIN, WD-GLT's average time is as high as $105.6\times$. Conversely, \ourmethod takes only $1.63-4\times$ the original dense training time to find the sparsest GLT. Notably, when it comes to finding a ticket with $s_\theta=49.7\%, s_g=36.9\%$ on Ogbn-Arxiv (Fig.~\ref{fig:time_arxiv}), FastGLT achieves 12.0x and 19.6x acceleration compared to UGS and WD-GLT, respectively.

\noindent \textbf{Obs.5. \ourmethod excels in obtaining more computationally efficient tickets. } Besides faster speed, \ourmethod also reduces computational load significantly. Across all GCN/GIN/GAT backbones, we achieve over 95\% MAC savings, surpassing UGS and WD-GLT by 5.4\% to 31.0\%.

\subsection{Ablation Study \& Sensitivity Analysis}
\label{sec:ablation}

\noindent \textbf{Validation of \textit{fast track}.} In this part, we validate the premise that \textit{one-shot tickets offer a fast track to winning tickets} from two aspects: convergence speed and optimal performance. Fig.~\ref{fig:abla+acc} (\textit{Right}) demonstrates that denoising from one-shot tickets not only successfully finds winning tickets, but also offers faster convergence. Tab.~\ref{tab:fast_track} compares the maximum $s_\theta$ and $s_g$ when starting from randomly initialized tickets versus one-shot tickets. Notably, denoising from random tickets results in up to a 28.94\% drop in weight sparsity and a 24.17\% drop in graph sparsity, highlighting the necessity of using one-shot tickets as a fast track.

\noindent\textbf{Effects of denoising interval $\Delta T$.} We set graph/weight sparsity at 40\%/80\%, and explore \ourmethod's performances with $\Delta T$ values $\{3,5,10,20,30,50\}$ on Ogbn-Arxiv+ResGCN and Citeseer+GAT. From Fig.~\ref{fig:abla+acc} (\textit{Left}), we observe that \ourmethod's sensitivity to $\Delta T$ is minimal, with a maximum accuracy variance of only 1.35\% on Citeseer and 1.74\% on Ogbn-Arxiv. The remaining parameter sensitivity analysis is provided in the Appendix.

\section{Related Work}


\subsubsection{Lottery Ticket Hypothesis} 
Current LTH methods are categorized into \textit{dense-to-sparse} and \textit{sparse-to-sparse} methods, based on the initial network's over-parameterization. The former starts with a dense network, progressively pruning to the target sparsity~\cite{chen2021gans,chen2021data,ding2021audio}, while the latter 
 starts with a randomly initialized sparse neural network and dynamically modifies its topology during a single training run through methods like pruning-and-growing~\cite{mocanu2018scalable,wang2020picking,liu2021we}. 
It's noteworthy that our gradual denoise mechanism bears a slight resemblance to the topology-adjusting operations of sparse-to-sparse methods. However, our work differs significantly in at least two ways: (1) Current sparse-to-sparse methods focus solely on weight pruning and aren't extendable to joint pruning. In contrast, our denoising mechanism is specifically devised based on observations of weight and graph characteristics; (2) Unlike sparse-to-sparse methods that initiate with a randomly sparse network, we start with one-shot pruning as a fast track to winning tickets, substantiated by extensive ablation studies proving its efficacy.

\subsubsection{Graph Lottery Ticket} UGS~\cite{chen2021unified} first integrated LTH's concept into GLT research, combining \emph{graph sparsification} with \emph{GNN model compression}. Recent endeavors include GEBT~\cite{you2022early}, which revealed the existence of graph early-bird tickets. WD-GLT~\cite{hui2022rethinking} enhanced graph pruning through an auxiliary loss function, and DGLT~\cite{wang2023searching} firstly introduced the concept of dual lottery tickets~\cite{bai2022dual} to GLT paradigm. However, all these methods fall into the trap of iterative pruning, making the acquisition of GLTs resource-intensive. Conversely, leveraging the one-shot pruning as a fast track, we bypass the extensive computational demands of IMP, acquiring winning tickets much more rapidly.

\section{Conclusions}
In this work, we propose an effective method termed one-shot pruning and denoising framework toward fast track graph lottery tickets (\ourmethod), which utilizes one-shot tickets as a \textit{fast track} and denoises them to acquire sparse but performant tickets. Our work reevaluates the relationship between one-shot and IMP tickets, hypothesizing and validating that one-shot tickets can be rapidly denoised to obtain subgraphs/subnetworks that are sparser and perform comparably to IMP-based tickets. This paradigm achieves a significantly faster efficiency in finding GLTs ($1.7-44\times$ speedup) compared to previous SOTA methods, offering new insights into how to more rapidly and effectively discover GLTs.

\bibliography{aaai25}

\section*{Reproducibility Checklist}

Unless specified otherwise, please answer “yes” to each question if the relevant information is described either in the paper itself or in a technical appendix with an explicit reference from the main paper. If you wish to explain an answer further, please do so in a section titled “Reproducibility Checklist” at the end of the technical appendix.

\subsection*{This paper:}

\begin{enumerate}
    \item Includes a conceptual outline and/or pseudocode description of AI methods introduced (yes)
    \item Clearly delineates statements that are opinions, hypothesis, and speculation from objective facts and results (yes)
    \item Provides well marked pedagogical references for less-familiar readers to gain background necessary to replicate the paper (yes)
\end{enumerate}

\subsection*{Does this paper make theoretical contributions? (no)}

If yes, please complete the list below.

\begin{enumerate}
    \item All assumptions and restrictions are stated clearly and formally. (yes/partial/no)
    \item All novel claims are stated formally (e.g., in theorem statements). (yes/partial/no)
    \item Proofs of all novel claims are included. (yes/partial/no)
    \item Proof sketches or intuitions are given for complex and/or novel results. (yes/partial/no)
    \item Appropriate citations to theoretical tools used are given. (yes/partial/no)
    \item All theoretical claims are demonstrated empirically to hold. (yes/partial/no/NA)
    \item All experimental code used to eliminate or disprove claims is included. (yes/no/NA)
\end{enumerate}

\subsection*{Does this paper rely on one or more datasets? (yes)}

If yes, please complete the list below.

\begin{enumerate}
    \item A motivation is given for why the experiments are conducted on the selected datasets (yes/)
    \item All novel datasets introduced in this paper are included in a data appendix. (NA)
    \item All novel datasets introduced in this paper will be made publicly available upon publication of the paper with a license that allows free usage for research purposes. (NA)
    \item All datasets drawn from the existing literature (potentially including authors’ own previously published work) are accompanied by appropriate citations. (yes)
    \item All datasets drawn from the existing literature (potentially including authors’ own previously published work) are publicly available. (yes)
    \item All datasets that are not publicly available are described in detail, with explanation why publicly available alternatives are not scientifically satisficing. (NA)
\end{enumerate}

\subsection*{Does this paper include computational experiments? (yes)}

If yes, please complete the list below.

\begin{enumerate}
    \item Any code required for pre-processing data is included in the appendix. (yes)
    \item All source code required for conducting and analyzing the experiments is included in a code appendix. (yes)
    \item All source code required for conducting and analyzing the experiments will be made publicly available upon publication of the paper with a license that allows free usage for research purposes. (yes)
    \item All source code implementing new methods have comments detailing the implementation, with references to the paper where each step comes from (yes)
    \item If an algorithm depends on randomness, then the method used for setting seeds is described in a way sufficient to allow replication of results. (NA)
    \item This paper specifies the computing infrastructure used for running experiments (hardware and software), including GPU/CPU models; amount of memory; operating system; names and versions of relevant software libraries and frameworks. (yes)
    \item This paper formally describes evaluation metrics used and explains the motivation for choosing these metrics. (yes)
    \item This paper states the number of algorithm runs used to compute each reported result. (yes)
    \item Analysis of experiments goes beyond single-dimensional summaries of performance (e.g., average; median) to include measures of variation, confidence, or other distributional information. (yes)
    \item The significance of any improvement or decrease in performance is judged using appropriate statistical tests (e.g., Wilcoxon signed-rank). (partial)
    \item This paper lists all final (hyper-)parameters used for each model/algorithm in the paper’s experiments. (yes)
    \item This paper states the number and range of values tried per (hyper-)parameter during development of the paper, along with the criterion used for selecting the final parameter setting. (yes)
\end{enumerate}

\clearpage
\appendix

\section{Algorithm Workflow}

We conclude the overall workflow of our \ourmethod in Algo.~\ref{alg:algo}.

\begin{algorithm}[!h]
\caption{Algorithm workflow of \ourmethod}
\label{alg:algo}
\textbf{Input}:$\mathcal{G}=(\mathbf{A},\mathbf{X})$, GNN model $f(\mathcal{G}, \mathbf{\Theta}_0)$, GNN's initialization $\mathbf{\Theta}_0$, target sparsity $s_g^{\text{tgt}}$ for graphs and $s_\theta^{\text{tgt}}$ for weights, denoising interval $\Delta T$, learning rate $\eta$.\\
\textbf{Output}:GLT\;$f\left( \{\mathbf{M}_g\odot \mathbf{A}, \mathbf{X}\}, \mathbf{M}_\theta \odot \mathbf{\Theta} \right)$
\begin{algorithmic}[1] 

\FOR{iteration $i = 1$ \TO $E$}
    \STATE Forward $f_{sub}\left( \{\boldsymbol{m}_g \odot \mathbf{A}, \mathbf{X} \}, \boldsymbol{m}_\theta \odot \mathbf{\Theta}\right)$ to compute the loss in Eq.~\ref{eq:loss}.
    \STATE Backpropagate to update $\mathbf{\Theta}_{i+1} \leftarrow \mathbf{\Theta}_{i} - \eta \nabla_{\mathbf{\Theta}}\mathcal{L}_{\text{os}}$.
    \STATE Update $\boldsymbol{m}_g^{i+1}, \boldsymbol{m}_\theta^{i+1}  \leftarrow \boldsymbol{m}_g^{i} - \eta \nabla_{\boldsymbol{m}_g^{i}}\mathcal{L}_{\text{os}}, \; \boldsymbol{m}_\theta^{i} - \eta \nabla_{\boldsymbol{m}_\theta^{i}}\mathcal{L}_{\text{os}}$.
\ENDFOR

\STATE Compute intermediate sparsity $s_g^{\text{inm}} \leftarrow \mathbf{\Upsilon}(s_g^{\text{tgt}}), \; s_\theta^{\text{inm}} \leftarrow \mathbf{\Upsilon}(s_\theta^{\text{tgt}})$.

\STATE \textbf{// One-shot Tickets.}
\STATE Set $s_g^{\text{inm}}\%$ of the lowest magnitude values in $\boldsymbol{m}_g^{E}$ to 0 and others to 1, then obtain one-shot mask $\mathbf{M}^{\circledcirc}_g$.
\STATE Set $s_\theta^{\text{inm}}\%$ of the lowest magnitude values in $\boldsymbol{m}_\theta^{E}$ to 0 and others to 1, then obtain one-shot mask $\mathbf{M}^{\circledcirc}_\theta$.

\STATE \textbf{// Gradual Denoising Procedure.}
\STATE Set $\mathbf{M}_g^{(1)} \leftarrow \mathbf{M}^{\circledcirc}_g$, $\mathbf{M}_\theta^{(1)} \leftarrow \mathbf{M}^{\circledcirc}_\theta$.

\FOR{iteration $d = 1$ \TO $D$}
    \STATE Compute interval index $\mu \leftarrow \lceil d / \Delta T \rceil$.
    \STATE Forward $f\left(  \{\boldsymbol{m}_g \odot \mathbf{M}^{(\mu)}_g \odot \mathbf{A}, \mathbf{X}\},  \mathbf{M}^{(\mu)}_\theta \odot \mathbf{\Theta} \right)$ to compute the $\mathcal{L}_{os}$.
    \STATE Update $\mathbf{\Theta}$ and $\boldsymbol{m}_g$ accordingly.
    \IF{$\mu = \lceil d / \Delta T \rceil$}
        \STATE Identify $\mathbf{M}_\theta^{(\text{noisy})}$ and $\mathbf{M}_g^{(\text{noisy})}$ according to Eq.~\ref{eq:noisy}.
        \STATE Identify $\mathbf{M}_\theta^{(\text{potential})}$ and $\mathbf{M}_g^{(\text{potential})}$ according to Eq.~\ref{eq:potential}.
        \STATE Compute $\mathbf{M}^{(\mu+1)}_g$ and $\mathbf{M}^{(\mu+1)}_\theta$ according to Eq.~\ref{eq:update}.
    \ENDIF
\ENDFOR

\end{algorithmic}
\end{algorithm}

\section{Experimental Settings}
\subsection{Hyperparameter Configuration}
We conclude the detailed hyperparameter configuration in Tab.~\ref{tab:parameter}.

\begin{table}[h]
  \centering
  \renewcommand\arraystretch{0.50}
  \caption{Hyper-parameter configurations.}
  \scriptsize
    \begin{tabular}{c|ccc ccc}
    \toprule
    \multicolumn{7}{c}{\centering \scriptsize Computing Infrastructures: NVIDIA Tesla V100} \\
    \midrule
     \multirow{2}{*}{Param} & \multicolumn{3}{c}{Cora}  & \multicolumn{3}{c}{Citeseer} \\
    
     & GCN & GIN & GAT & GCN & GIN & GAT \\
    \cmidrule(r){1-1} \cmidrule(r){2-4}\cmidrule(r){5-7} 
    $E$ & 30 & 30 & 30 & 50 & 50 & 50 \\
    $D$ & 400 & 500 & 600 & 400 & 500 & 400 \\
    lr & 0.001 & 0.001 & 0.001 & 0.001 & 0.001 & 0.001 \\
    $\Delta T$ & 10 & 10 & 10 & 10 & 10 & 3 \\
    \midrule
     \multirow{2}{*}{Param} & \multicolumn{3}{c}{PubMed} & Arxiv & Proteins & Collab \\
     
    & GCN & GIN & GAT & \multicolumn{3}{c}{ResGCN}\\
    \cmidrule(r){1-1} \cmidrule(r){2-4}\cmidrule(r){5-7} 
     $E$ & 50 & 50 & 30 &30 & 20 & 50 \\
     $D$ & 400 & 500 & 400&400 & 200 & 400\\
     lr & 0.001 & 0.001 & 0.001&0.001& 0.001&0.001 \\
     $\Delta T$ & 20 & 10 & 10 & 3& 3& 3 \\
     
    \bottomrule
    \end{tabular}%
  \label{tab:parameter}%
\end{table}%

\section{Additional Experimental Results}
\subsection{Additional Results on OGB Graphs}
As showcased in Tab.~\ref{tab:large_graph_weight}, 
\ourmethod consistently identifies GLTs with weight sparsity over 70\% across three OGBN datasets, surpassing other methods which generally fall below 60\%.

\begin{table}[!t]\scriptsize
\centering
\caption{Results on 3 large-scale OGB graphs. 
Each entry denotes the extreme sparsity that a certain method is capable of achieving. $\pm$ corresponds to the standard deviation over 5 trials. ``N/A'' means GLT cannot be found.}
\setlength{\tabcolsep}{3.40pt}
\begin{tabular}{lccc}
\toprule
Dataset & Ogbn-Arxiv & Ogbn-Proteins & Ogbl-Collab  \\ 
\hline
\multicolumn{4}{c}{\textbf{Weight Sparsity}}  \\ \hline
Random Pruning & $28.72\pm 2.38$ & $17.88\pm 3.66$  & $6.14\pm 1.54$ \\
UGS\citep{chen2021unified}   & $40.07\pm 0.16$ & $33.60\pm0.45 $  & $38.32\pm0.24$\\
WD-GLT\citep{hui2022rethinking} & $63.23\pm0.11$ & $47.23\pm0.18$ & $59.06\pm1.37$ \\
\ourmethod (Ours) & {{$73.25 \pm 0.18$}} & {{$73.01\pm 0.34$}} &  {{$71.66 \pm 0.45$}}  \\
\bottomrule
\end{tabular}
\label{tab:large_graph_weight}
\end{table}

\subsection{Ablation Study Results}
\begin{figure}[h]
\centering
\includegraphics[width=1\columnwidth]{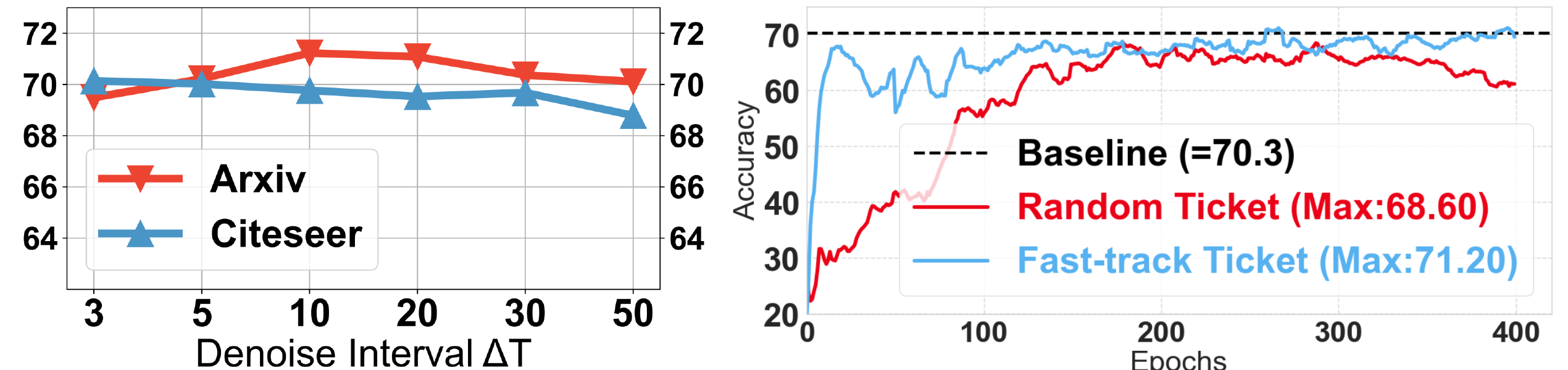}
\caption{(\textbf{Left}) Ablation study on $\Delta T$. We vary $\Delta T \in \{3,5,10,20,30,50\}$ on Ogbn-Arxiv+ResGCN and Citeseer+GAT with fixed sparsity $\{ s_\theta = 80\%, s_g = 40\% \}$; (\textbf{Right}) Test accuracy curves showcasing the denoising process from randomly initialized tickets and one-shot tickets on Citeseer+GCN with $\{s_g=30\%,s_\theta=99\%\}$. }
\label{fig:abla+acc}
\end{figure}

We conduct ablation studies from three distinct perspectives: one-shot pruning(Tab. ~\ref{tab:fast_track}), denoising interval $\Delta T$(Fig. ~\ref{fig:abla+acc}), and denoising scheduler $\mathbf{\Upsilon}$(Tab. ~\ref{tab:decay}). Specifically, we have three observations:
\begin{itemize}[leftmargin=*]
\item As shown in Tab.~\ref{tab:fast_track}, one-shot tickets effectively provide a \textit{fast track} to winning tickets.
\item As shown in Fig.~\ref{fig:abla+acc}, the denoising interval $\Delta T$ does not have a significant effect on the performance of \ourmethod.
\item As shown in Tab.~\ref{tab:decay}, \ourmethod is not sensitive to denoising scheduler $\mathbf{\Upsilon}$.
\end{itemize}

\begin{table}[h]
  \centering
  \renewcommand\arraystretch{0.50}
  \small
\caption{Ablation study on denoising from randomly initialized tickets and one-shot tickets on Citeseer with GCN/GIN/GAT.}\label{tab:fast_track}
\begin{tabular}{lccc}
\toprule
Backbone & GCN & GIN & GAT  \\ 
\midrule
& \multicolumn{3}{c}{{Weight Sparsity} }  \\ 
\midrule
Random Ticket & $94.22$ &  $89.38$  & $70.26$ \\
One-shot Ticket&  $99.27$ &  $98.33$  &  $99.20$ \\
\midrule
& \multicolumn{3}{c}{{Graph Sparsity} }  \\ 
\midrule
Random Ticket& $23.16$ & $20.89$  & $56.06$ \\
One-shot Ticket&  $34.17$ &  $45.06$  &  $68.97$ \\
\bottomrule
\end{tabular}
\end{table}%

\begin{table}[h]
  \centering
  \renewcommand\arraystretch{0.50}
  \small
  \caption{Ablation study on decay factor $\kappa$. We report performances of GCN on Cora ($s_g=30\%,s_\theta=90\%$) and Citeseer ($s_g=40\%,s_\theta=90\%$), with $\kappa \in \{1,2,3\}$.}\label{tab:decay}
\begin{tabular}{lccc}
\toprule
(Cora) & GCN & GIN & GAT  \\ 
\midrule
$\kappa=1$  & $80.09_{\pm 0.28}$ &  $78.26_{\pm0.44}$  & $80.77_{\pm0.18}$ \\
$\kappa=2$ &  $79.84_{\pm0.23}$ &  $78.04_{\pm0.09}$  &  $79.80_{\pm0.11}$ \\
$\kappa=3$ &  $79.86_{\pm0.21}$ &  $77.71_{\pm0.58}$  &  $80.54_{\pm0.37}$\\
\midrule
(Citeseer) & GCN & GIN & GAT  \\ 
\midrule
$\kappa=1$  & $69.26_{\pm0.04}$&  $70.08_{\pm0.31}$  & $69.88_{\pm0.12}$ \\
$\kappa=2$ &  $69.23_{\pm0.13}$ &  $69.16_{\pm0.39}$  &  $69.82_{\pm0.16}$ \\
$\kappa=3$ 
& $69.06_{\pm0.71}$ 
& $68.14_{\pm0.15}$  
& $68.93_{\pm0.45}$\\
\bottomrule
\end{tabular}
\end{table}%

\subsection{Parameter Sensitivity Analysis}
\noindent\textbf{Effects of denoising interval $\Delta T$.} We set graph/weight sparsity at 40\%/80\%, and explore \ourmethod's performances with $\Delta T$ values $\{3,5,10,20,30,50\}$ on Ogbn-Arxiv+ResGCN and Citeseer+GAT. From Fig.~\ref{fig:abla+acc} (\textit{Left}), we observe that \ourmethod's sensitivity to $\Delta T$ is minimal, with a maximum accuracy variance of only 1.35\% on Citeseer and 1.74\% on Ogbn-Arxiv. 

\noindent \textbf{Effects of denoising scheduler $\mathbf{\Upsilon}$.} We evaluate the accuracy fluctuations across various GNN backbones on Cora/Citeseer when varying the decay factor $\kappa$ of Inverse Power at $\{ 1,2,3 \}$. The magnitude of $\kappa$ is inversely related to the decay rate. Tab.~\ref{tab:decay} shows that linear decay ($\kappa=1$) outperforms the other two decay rates, and we therefore uniformly apply $\kappa=1$ in all experiments. Still, the maximum performance variance on Cora/Citeseer is less than 0.55\%/1.94\% respectively, indicating \ourmethod's low sensitivity to the denoising scheduler.

\subsection{Comparison of One-shot Pruning and FastGLT}

We have included a straightforward one-shot pruning method as an additional baseline for comparison. The experimental results, as shown in Table~\ref{tab:one_shot_comparison}, demonstrate that FastGLT outperforms the basic one-shot approach, highlighting the benefits of our denoising framework in refining one-shot tickets to achieve higher accuracy and greater sparsity.

\begin{table}[h]
  \centering
  \renewcommand\arraystretch{0.50}
  \caption{Accuracy(Acc) / Extreme Sparsity(ES) Comparison of One-shot Pruning and FastGLT on Different Datasets and Backbones.}
  \scriptsize
    \begin{tabular}{c|ccc ccc}
    \toprule
    \multicolumn{7}{c}{\centering \scriptsize Comparison of One-shot Pruning and FastGLT with GCN} \\
    \midrule
     \multirow{2}{*}{Method} & \multicolumn{2}{c}{Cora}  & \multicolumn{2}{c}{Citeseer} & \multicolumn{2}{c}{Pubmed} \\
     & Acc(\%) & ES(\%) & Acc(\%) & ES(\%) & Acc(\%) & ES(\%) \\
    \cmidrule(r){1-1} \cmidrule(r){2-3}\cmidrule(r){4-5} \cmidrule(r){6-7} 
    Baseline & 80.25 & - & 70.51 & - & 78.80 & - \\
    One-Shot & 80.09 & 15.00 & 70.65 & 25.00 & 79.06 & 25.00 \\
    FastGLT & \textbf{80.33} & \textbf{35.00} & \textbf{79.11 } & \textbf{35.00} & \textbf{79.11} & \textbf{40.00} \\
    \midrule
    \multicolumn{7}{c}{\centering \scriptsize Comparison of One-shot Pruning and FastGLT with GAT} \\
    \midrule
     \multirow{2}{*}{Method} & \multicolumn{2}{c}{Cora}  & \multicolumn{2}{c}{Citeseer} & \multicolumn{2}{c}{Pubmed} \\
     & Acc & ES & Acc & ES & Acc & ES \\
    \cmidrule(r){1-1} \cmidrule(r){2-3}\cmidrule(r){4-5} \cmidrule(r){6-7} 
    Baseline & 79.95 & - & 69.12 & - & 78.35 & - \\
    One-Shot & 79.82 & 45.00 & 69.48 & 50.00 & 78.39 & 40.00 \\
    FastGLT & \textbf{80.17} & \textbf{75.00} & \textbf{69.30} & \textbf{82.50} & \textbf{78.56} & \textbf{70.00} \\
    \bottomrule
    \end{tabular}%
  \label{tab:one_shot_comparison}%
\end{table}%

\subsection{Extreme Sparsity of Winning Tickets}

We report the extreme sparsity of Winning Tickets to enhance clarity in our results, as shown in Table~\ref{tab:extreme_sparsity}. For One-Shot Pruning and FastGLT, we iteratively search through each sparsity level in an arithmetic sequence until a lottery ticket can no longer be found. FastGLT consistently identifies winning tickets with higher sparsity compared to the baseline, highlighting its superiority.

\begin{table}[h]
  \centering
  \renewcommand\arraystretch{0.50}
  \caption{Detailed Extreme Sparsity of Winning Tickets}
  \scriptsize
    \begin{tabular}{c|cc cc}
    \toprule
     \multirow{2}{*}{Method} & \multicolumn{2}{c}{GCN}  & \multicolumn{2}{c}{GAT} \\
     & Cora & Pubmed & Cora & Pubmed  \\
    \cmidrule(r){1-1} \cmidrule(r){2-3}\cmidrule(r){4-5} 
    One-Shot & 15.00\% & 25.00\% & 45.00\% & 40.00\% \\
    WD-GLT & 18.52\% & 34.65\% & 60.12\% & 62.45\% \\
    FastGLT  & \textbf{35.00\%}  & \textbf{40.00\%}      & \textbf{75.00\%}    & \textbf{70.00\%}  \\
    \bottomrule
    \end{tabular}%
  \label{tab:extreme_sparsity}%
\end{table}%

\section{Practical Usage of GLT}
Graph Lottery Ticket has been confirmed to have broad practical applications. We provide examples as follows:

\begin{itemize}
    \item \textbf{Acceleration:} Our experiments on multiple datasets and GNNs show that FastGLT achieves an average inference speedup of 1.48x. Moreover, GLT accelerates Neural Architecture Search by applying graph sparsification combined with architecture-aware edge deletion~\cite{Xie_2024}.
    \item \textbf{Transferability:} The winning ticket can be transferred across different datasets \& GNNs, eliminating the need for retraining.
    \item \textbf{Robustness:} GLT aids in detecting redundant and poisoned edges, enhancing input perturbation robustness.
    \item \textbf{Federated Learning(FL) Data Compression:} In FL, GLT obtains sparse structures from local GNNs, reducing the parameter load sent to the central server.
\end{itemize}

\section{Dataset and Backbone}
\subsection{Graph datasets statistics}

We conclude the dataset statistics in Tab.~\ref{tab:dataset_part1}
 and Tab.~\ref{tab:dataset_part2}.
 
\begin{table}[h]
\centering
\caption{Graph datasets statistics (Part 1).}
\label{tab:dataset_part1}
\begin{tabular}{c  |c |c |c }
\toprule
Dataset   &  Nodes & Edges & Avg. Degree \\ 
\midrule
Cora  & 2,708 &5,429 & 3.88 \\ 
Citeseer & 3,327 & 4,732 &1.10 \\ 
PubMed& 19,717 & 44,338 & 8.00 \\ \midrule

Ogbn-ArXiv  & 169,343 & 1,166,243 & 13.77 \\
Ogbn-Proteins  & 132,534 & 39,561,252 & 597.00 \\
Ogbl-Collab & 235,868 & 1,285,465 & 10.90 \\
\bottomrule
\end{tabular}
\vspace{-2mm}
\end{table}

\begin{table}[h]
\centering
\vspace{-4mm}
\caption{Graph datasets statistics (Part 2).}
\label{tab:dataset_part2}
\begin{tabular}{c  |c |c |c }
\toprule
Dataset   &  Features & Classes & Metric  \\ 
\midrule
Cora  & 1,433  & 7 & Accuracy\\ 
Citeseer & 3,703 & 6 & Accuracy \\ 
PubMed& 500  & 3 &  Accuracy \\ \midrule

Ogbn-ArXiv  & 128 & 40 & Accuracy \\
Ogbn-Proteins  & 8 & 2 & ROC-AUC \\
Ogbl-Collab & 128 & 2 & Hits@50 \\
\bottomrule
\end{tabular}
\vspace{-2mm}
\end{table}

\subsection{Performance Metrics}
Accuracy represents the ratio of correctly predicted outcomes to the total predictions made. The ROC-AUC (Receiver Operating Characteristic-Area Under the Curve) value quantifies the probability that a randomly selected positive example will have a higher rank than a randomly selected negative example. Hit@50 denotes the proportion of correctly predicted edges among the top 50 candidate edges.

\subsection{Efficiency Metrics}\label{app:eff}
To evaluate the efficiency of sparse graphs generated by different sparsifiers, we employ two key metrics: MACs (Multiply-Accumulate Operations) and GPU inference latency (measured in milliseconds). MACs are a theoretical indicator of the model's inference speed, based on FLOPs (Floating Point Operations Per Second). Although {\fontfamily{lmtt}\selectfont \textbf{SpMM}} is theoretically faster than {\fontfamily{lmtt}\selectfont \textbf{MatMul}} according to MACs/FLOPs, this advantage is not always evident in practice due to {\fontfamily{lmtt}\selectfont \textbf{SpMM}}'s irregular memory access pattern. To better understand the practical performance of our approach, we also measure the GPU latency in milliseconds.

\subsection{Train-val-test Splitting of Datasets.} To rigorously verify the effectiveness of \ourmethod, we unify the dataset-splitting strategy across all GNN backbones and baselines. As for node classification tasks of small- and medium-size datasets, we utilize 140(Cora), 120 (Citeseer) and 60 (PubMed) labeled data for training, 500 nodes for validation and 1000 nodes for testing.  The data splits for Ogbn-ArXiv, Ogbn-Proteins, and Ogbl-Collab were provided by the benchmark~\cite{hu2020open}. Specifically, for Ogbn-ArXiv, we train on papers published until 2017, validate on papers from 2018 and test on those published since 2019. For Ogbn-Proteins, protein nodes were segregated into training, validation, and test sets based on their species of origin. For Ogbl-Collab, we employed collaborations until 2017 as training edges, those in 2018 as validation edges, and those in 2019 as test edges.

\subsection{More Details about Backbones} As for small- and medium-scale datasets Cora, Citeseer and PubMed, we choose the two-layer GCN/GIN/GAT networks with 512 hidden units to conduct all our experiments. As for large-scale datasets Ogbn-ArXiv, Ogbn-Proteins and Ogbl-Collab, we use the ResGCN with 28 GCN layers to conduct all our experiments.

For comparison with state-of-the-art GLT methods, we choose UGS  and TGLT, which are the most efficient GLT methods to our best knowledge. 

\begin{itemize}
    \item \textbf{UGS}~\cite{chen2021unified}: We utilize the official implementation from the authors. Notably, UGS was originally designed for joint pruning of model parameters and edges. Specifically, it sets separate pruning parameters for parameters and edges, namely the weight pruning ratio $p_\theta$ and the graph pruning ratio $p_g$. In each iteration, a corresponding proportion of parameters/edges is pruned. For a fairer comparison, we set $p_\theta=0\%$ and $p_g \in \{5\%,10\%\}$ to get the results of all sparsity granularity.
    \item \textbf{WD-GLT}~\cite{hui2022rethinking}: WD-GLT inherits the iterative magnitude pruning paradigm from UGS, so we also set $p_\theta=0\%$ and $p_g \in \{5\%,10\%\}$ across all datasets and backbones. The perturbation ratio $\alpha$ is tuned among $\{0,1\}$. Since no official implementation is provided, we carefully reproduced the results according to the original paper.
\end{itemize}

\end{document}